\documentclass[sigconf]{acmart}

\usepackage{subfigure}
\usepackage{multirow}
\usepackage{graphicx}
\usepackage{enumitem}
\usepackage{cleveref}
\usepackage{fancyhdr}

\begin{document}


\setcopyright{acmcopyright}
\copyrightyear{2020}
\acmYear{2020}
\acmDOI{xx.xxxx/xxxxxxxxx.xxxxxxx}

\acmConference[epiDAMIK 2020]{3rd epiDAMIK ACM SIGKDD International Workshop on Epidemiology meets Data Mining and Knowledge Discovery}{Aug 24, 2020}{San Diego, CA}
\acmBooktitle{epiDAMIK 2020: 3rd epiDAMIK ACM SIGKDD International Workshop on Epidemiology meets Data Mining and Knowledge Discovery}
\acmPrice{15.00}
\acmISBN{978-1-xxxx-XXXX-X}

\title{Neural Networks for Pulmonary Disease Diagnosis using Auditory and Demographic Information}

\author{Morteza Hosseini}
\affiliation{
  \institution{University of Maryland\\
  Baltimore County}}

\author{Haoran Ren}
\affiliation{
  \institution{University of Maryland\\
  Baltimore County}}

\author{Hasib-Al Rashid}
\affiliation{
  \institution{University of Maryland\\
  Baltimore County}}

\author{Arnab Neelim Mazumder}
\affiliation{
  \institution{University of Maryland \\
  Baltimore County}}

\author{Bharat Prakash}
\affiliation{%
  \institution{University of Maryland \\
  Baltimore County}}

\author{Tinoosh Mohsenin}
\affiliation{
  \institution{University of Maryland\\
  Baltimore County}}


\begin{abstract}
Pulmonary diseases impact millions of lives globally and annually. The recent outbreak of the pandemic of the COVID-19, a novel pulmonary infection, has more than ever brought the attention of the research community to the machine-aided diagnosis of respiratory problems. This paper is thus an effort to exploit machine learning for classification of respiratory problems and proposes a framework that employs as much correlated information (auditory and demographic information in this work) as a dataset provides to increase the sensitivity and specificity of a diagnosing system. First, we use deep convolutional neural networks (DCNNs) to process and classify a publicly released pulmonary auditory dataset, and then we take advantage of the existing demographic information within the dataset and show that the accuracy of the pulmonary classification increases by 5\% when trained on the auditory information in conjunction with the demographic information. Since the demographic data can be extracted using computer vision, we suggest using another parallel DCNN to estimate the demographic information of the subject under test visioned by the processing computer. Lastly, as a proposition to bring the healthcare system to users' fingertips, 
we measure deployment characteristics of the auditory DCNN model onto processing components of an NVIDIA TX2 development board.
\end{abstract}






\maketitle

\pagestyle{fancy}
\rhead{\textcolor{black}{\textit{\textbf{\small ACM SIGKDD, epiDAMIK, August 2020, San Diego, CA}}}}

\section{Introduction}

In 2016, pulmonary diseases were among the top 10 causes of death: ranked 1 for low-income and ranked 5 for high-income countries \cite{who2016}. Recently, with the outbreak of the COVID-19 as a novel pulmonary infection, a tremendous amount of attention has been directed to control the pandemic crisis about which extreme measures are taken by countries to diagnose the infected patients. Measures such as extensive testing and early-stage diagnosis help to locate and contain the infection, and are reportedly the most effective preventive actions to control the contagion in a pandemic.

\begin{figure*}[htb]
\centering
\includegraphics[width=7in]{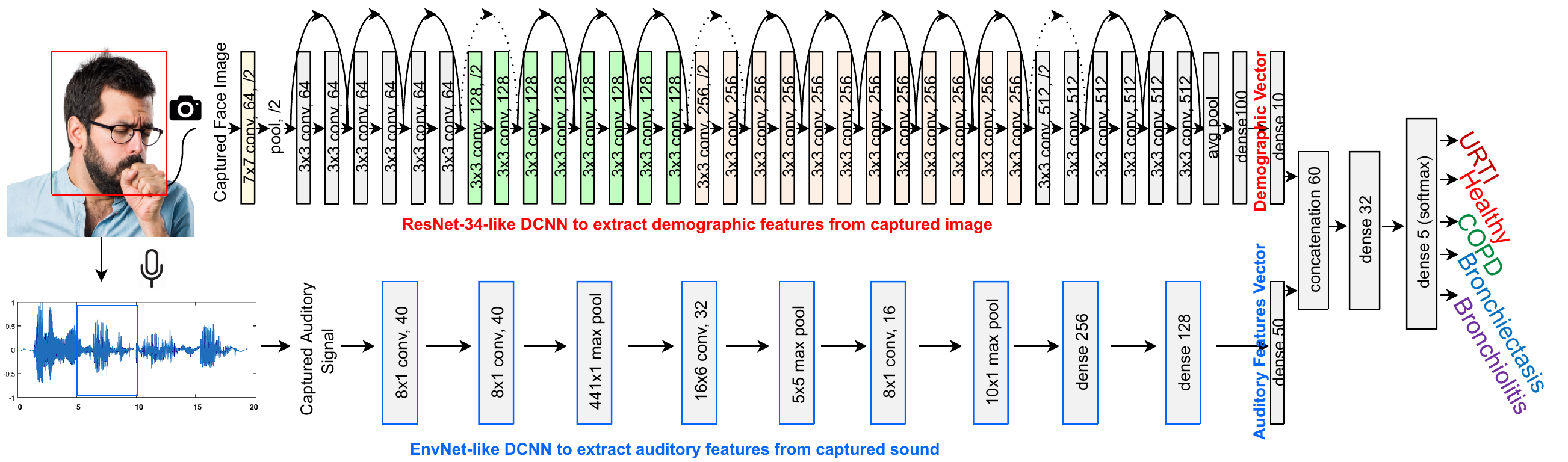}
\caption{The proposed framework to classify respiratory problem has two DCNN components that process data from a user under test. Part of the information is auditory, such as the audio sound recorded from a medical electronic device like a microphone or a stethoscope, and part of that information is the demographic information, such as age, gender, and ethnicity, that can either be estimated using a computer vision algorithm or inserted manually. The framework is flexible and scalable in the sense that it can incorporate new sensors easily, allowing the system to be tailored to a variety of kinds of situations, such as in-home consultations, clinical visits, or even symptom detection in public milieus using non-contact sensors.}
\label{fig:proposed}
\end{figure*}

Pulmonary problems encompass a wide range of chronic and infectious diseases, and because of the common organ, lung, that they affect, they develop respiratory symptoms whose auditory signals recorded from various medical devices are among the first to be scrutinized by a medical expert. As an example, COVID-19 develops symptoms such as dry cough, fever, fatigue, dyspnea, and shortness of breath that vary in severity at different stages of the disease progression, and correlate with certain ethnicity, gender, and age groups differently \cite{liu2020clinical}. More than 70\% of the confirmed COVID-19 patients have reported fever in tandem with a dry cough \cite{zhao2020incidence}. Meanwhile, clinical case records indicate that the young population is less likely to develop COVID-19 relevant symptoms, contrary to the elderly that is the most vulnerable group \cite{lee2020children}. 

Traditionally, when someone feels symptoms, they either call a doctor or have themselves seen/scrutinized by medical experts at walk-in clinics, where extensive use of vital signs, visual and auditory information are applied to make diagnostic decisions. Such practice during a pandemic or in remote locations is unsuitable/impractical as a result of the limited capacity of existing facilities and human resources at health centers, and, ironically, can expedite spreading the infection. On the other hand, calls are made by governments/organizations during the pandemic for people to stay at home that, by itself, has caused a state of confusion and has made another barrier. Thus, early-stage and clinic independent machine assistance is critical for the initial diagnosis of the disease and/or for evaluating/assessing its severity.


Our goal in this research is to allow machine learning algorithms running on general computing processors (e.g., those in cell-phones and tablets) to assess patients similar to what doctors do at triage and telemedicine, using passively recorded audio and/or video and self-declared information, to bring proactive healthcare to users’ fingertips and to estimate the urgency/necessity of whether they need to attend clinics and have themselves further examined with the use of more specialized test-kits or facilities. Our vision is to provide a detection framework that can provide early detection for anyone and anywhere. We develop our work on a publicly released respiratory sound database that includes both auditory and demographic information recorded from 126 subjects covering 7 pulmonary diseases including healthy condition, and with two sets of annotations. More specifically, since the respiratory sound dataset includes the two types of information per patient, we examine how the lack/existence of the demographic information impacts the total accuracy of the model. For further compilation, we develop another deep neural network that estimates demographic information including the age and the gender of the captured images and to correlate them with the auditory signals recorded from the subject under test in order to assess a higher sensitivity and specificity rate of diagnosis. The main contributions of this work include:
\begin{itemize}[leftmargin=0.25in]
 \item Statistically analyze the information in a public respiratory sound database, to justify extracting a reasonably balanced dataset out of it.
 \item Train a DCNN on the extracted auditory dataset without considering the demographic information.
 \item Train another DCNN model on a face images dataset annotated with age, gender, and ethnicity so to estimate/extract demographic information of a subject visioned by a computer.
 \item Train the auditory dataset in conjunction with the demographic information.
 \item Deploy the first DCNN model to TX2 embedded system and measure its implementation characteristics for CPU and GPU. 
\end{itemize}

\section{Related Work}
With the advancement of machine learning and deep learning algorithms, audio-based biomedical diagnosis and anomaly detection have recently become an active area of research. Some important aspects of audio-based diagnosis using deep learning include detection of sleep apnea, recognition of cough tone, and classification of heart sound, to name a few. Early research \cite{imran2020ai4covid} shows that machine learning (ML) tools on a limited unpublished dataset can distinguish solely between coughs from COVID-19 patients and those who are healthy or with upper-respiratory coughs with high accuracy of 96.8\%. \cite{Rash2009:End} introduces End to End convolutional neural networks for cough and dyspnea detection. Authors in \cite{amoh2016deep} used both DCNNs and recurrent neural networks (RNNs) to classify cough sound that they collected using chest-mounted sensors. Authors in \cite{nakano2019tracheal} used deep learning to detect sleep apnea.
Classification of heart sound into normal and abnormal classes was conducted in \cite{ryu2016classification} using DCNNs. Authors in \cite{aykanat2017classification} and \cite{perna2019deep} used DCNNs and RNNs to classify lung sounds respectively. Most of these works report high levels of accuracy on unpublished datasets that are accessible by the research community.
The 2017 International Conference on Biomedical Health Informatics (ICBHI) \cite{rocha2017alpha} issued a benchmark dataset of respiratory sound to facilitate researching on respiratory sound classification. Since then, researchers proposed various algorithms \cite{perna2018convolutional, liu2019detection, pham2020robust, demir2020convolutional} using different deep learning techniques to classify respiratory cycle anomalies such as the precise locations of wheezes and crackles within the cycle of each respiratory sound recording. 
In \cite{ma2019lungbrn} authors showed innovativeness by proposing a digital stethoscope to provide an immediate diagnosis of respiratory diseases. They developed a modified bi-ResNet architecture using STFT and wavelet feature extraction. Log quantized deep CNN-RNN based model for respiratory sound classification was proposed in \cite{acharya2020deep} for memory limited wearable devices. 

\begin{figure*}[htb]
\centering
\includegraphics[width=7in]{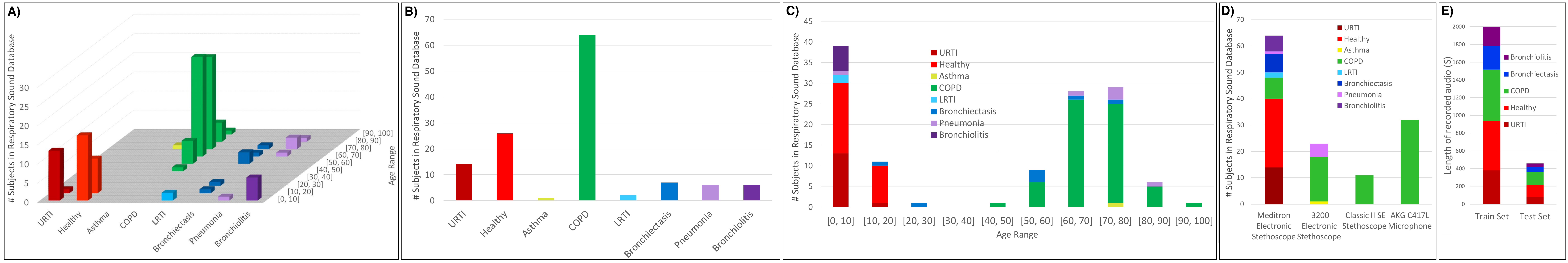}
\caption{Statistics of the respiratory sound database that contains auditory samples from 126 patients, A) a 2D histogram of 7 pulmonary classes with respect to 10 age groups, B) Break-down of the pulmonary classes in the dataset, C) Break-down of each pulmonary class with respect to age groups, D) Break-down of each pulmonary class with respect to the four recording medical devices, E) Our selection of 52 and 11 individual subjects for train and test datasets respectively that cover 5 pulmonary classes recorded with Meditron Electronic Stethoscope.}
\label{fig:ICHBI}
\end{figure*}

\section{Proposed Method}
The framework, depicted in Fig. \ref{fig:proposed}, leverages audio/video to extract necessary and medically relevant information and combines the extracted features with other inserted/self-declared patient data. The audio processing incorporates an ML approach such as a DCNN that extracts symptomatic features like
crackles and wheezes of lung sounds from a given window of recorded sound of a subject under test. At the video processing path, the captured RGB images are given to a ResNet-34 DCNN to process and estimate the user’s other demographic and symptomatic features such as age and gender. The extracted audio/video features along with the other relevant inserted data are concatenated towards the final layers and with the addition of a few more neural network layers or an ensemble of classifiers in the last layer, a probability vector of diagnosis for the user under test is reported. Both audio/video data can extend the scope of the clinical-reported symptoms to more diverse features that may be invisible to a human’s perception.
For example, when listened by a trained machine, the features extracted from the sound of a patient’s cough can include more useful features beyond terms like “dry” or “productive” that are commonly reported in clinical case records.

\subsection{Datasets}
\subsubsection{Respiratory Sound Database}
For the auditory dataset, we used a public respiratory sound database \cite{rocha2019open}, which includes 920 recordings acquired from 126 participants annotated with 8 types of respiratory conditions including URTI, Healthy, Asthma, COPD, LRTI, Bronchiectasis, Pneumonia, and Bronchiolitis. The recordings were collected using four types of medical equipment including AKG C417L Microphone, 3M Littmann Classic II SE Stethoscope, 3M Littmann 3200 Electronic Stethoscope, and Welch Allyn Meditron Master Elite Electronic Stethoscope. The duration of each recording range from 10 to 90 seconds mostly dominated with 20s samples. Fig. \ref{fig:ICHBI}-A, B, and C plot the distribution of the subjects with respect to their diagnosed disease and the age groups they impact, and Fig. \ref{fig:ICHBI}-D shows the contribution of each of the 4 medical devices for recordings from participants. Among the four recording devices, the Meditron Electronic Stethoscope is the only device that encompasses the 8 pulmonary conditions except for the Asthma, and is used for 63 out of the 126 participants. The recordings from the other 3 devices are majorly taken from COPD-diagnosed participants. By eliminating Asthma, Pneumonia, and LRTI that have little or no samples within the Meditron recordings, we extracted a random subset encompassing all 63 participants and split it into a semi-balanced train and a test set of 52 and 11 participants that include 5 types of pulmonary classes.  Fig. \ref{fig:ICHBI}-E shows a plot of the selected train/test dataset based on the total duration of each class. The database is meanwhile provided with demographic information of the 126 participants and another annotation that marks begin/end of respiratory cycles and the precise locations of events of crackles and wheezes per recording. Based on the second annotation, we counted the total number of respiratory cycles to estimate the slowest and average respiratory cycles within the dataset and to decide on a window size to cut the recordings into smaller frames. Table \ref{tab:selecteddataset} summarized the number of subjects, duration of recordings, and the number of respiratory cycles per pulmonary class within both train and test datasets.

\begin{table}
\caption{Selected semi-balanced Dataset out of the respiratory sound database. Meditron recordings from 61 patients that include 5 pulmonary classes were selected and the 20s sounds were chopped into overlapping frames of 5s. The total dataset of frames includes 1968 samples that were split into mutually-subject-exclusive between train (81\%) and test (19\%) subsets.}
\scriptsize
\setlength{\tabcolsep}{3.5pt}
\begin{center}
\begin{tabular}{c|c|c|c|c|c|c|c}
\hline
\multicolumn{2}{c|}{Selected Dataset} & URTI & Healthy & COPD & Bronchiec. & Bronchiol. & Total \\ \hline
\multirow{4}{*}{\begin{tabular}[c]{@{}c@{}}Train\\ Set\end{tabular}} & \# Subject & 12 & 24 & 6 & 5 & 5 & 52 \\
 & Duration (S) & 380 & 560 & 580 & 260 & 220 & 2000 \\
 & \# Respiratory Cycles & 207 & 257 & 406 & 88 & 141 & 1099 \\
 & \# Augmented Frames & 304 & 448 & 464 & 208 & 176 & 1600 \\ \hline
\multirow{4}{*}{\begin{tabular}[c]{@{}c@{}}Test\\ Set\end{tabular}} & \# Subject & 2 & 4 & 2 & 2 & 1 & 9 \\
 & Duration (S) & 80 & 140 & 140 & 60 & 40 & 460 \\
 & \# Respiratory Cycles & 26 & 48 & 119 & 17 & 16 & 226 \\
 & \# Augmented Frames & 64 & 112 & 112 & 48 & 32 & 368 \\ \hline
\end{tabular}
\end{center}
\label{tab:selecteddataset}
\end{table}

\subsubsection{Face Images Database}
UTKFace dataset \cite{zhifei2017cvpr} is a large-scale dataset consisting of over 20,000 face images with annotations of age (ranging from 0 to 116 years old) gender, and ethnicity. In \cite{karkkainen2019fairface}, the UTKFace dataset is trained on a ResNet-34 \cite{he2016deep}, and we reproduced the results of training over ResNet-34, and report the accuracy it gives for precise age as well as for the age group of a random split of 20\% test data.

\subsection{Data Pre-processing and Augmentation}
For data augmentation of the respiratory sound database, every recorded audio sample is cut into frames with a duration of 5s and with a stride of 1s, which means every two adjacent frames overlap a duration of 4s, and every 20s recorded sample results in 16 5s frames. Therefore the total 2000 seconds of the training dataset generates 1600 frames, and the total 460s testing data generates 368 frames of 5s samples. The choice of the 5s frames is inferred empirically by experiencing frames ranging from 1s to 10s. 

For the data augmentation of the UTKface images, we use common image augmentation techniques such as flipping, shifting and resizing the images within the dataset. 


\section{Experimental Setup}
We used a ResNet-34 DCNN \cite{he2016deep} for the UTKface RGB images of size 200$\times$200, and an EnvNet-like \cite{tokozume2017learning} DCNN for the respiratory sound frames of size 1$\times$220500. For the EnvNet-like DCNN, the input from the audio recordings is a one-dimensional vector where the size depends on the window selected for the framework. To best utilize the one-dimensional input, we use two one dimensional convolution layers to extract relevant features with a follow up of non-overlapping max-pooling operation to downsample the feature map. The subsequent layers include two-dimensional convolutional layers with max-pooling layers in between for efficient classification of the diseases. Finally, the fully connected layers summarize the required feature information and feed it to the extended model to generate a generalized output that classifies 5 types of pulmonary conditions as in our extracted dataset.


\subsection{Demographic Classification}
\subsubsection{ResNet-34 for UTKFace}\hfill\\
The classification accuracy of age and gender estimation of ResNet-34 is reported in Table \ref{tab:utkface}. Although the DCNN model does not precisely classify the age within the test dataset, it is able to classify the gender and estimate the age groups when the range of the groups expands. This is in correspondence to combining the auditory data with the age group, as conducted and reported in the next subsection where we combine the auditory information with the age group of the subjects, rather than the precise age of each participant. 


\begin{table}
\caption{ The classification accuracy and the model complexity of a ResNet-34 DCNN that extracts demographic information on the UTKface test dataset.}
\scriptsize
\setlength{\tabcolsep}{7.5pt}
\begin{center}
\begin{tabular}{c|cc|cccc}
\hline
\multicolumn{3}{c|}{DCNN characteristics} & \multicolumn{4}{c}{Test accuracy} \\ \hline
Model & \#params & FLOPs & Age±0 & Age±5 & \multicolumn{1}{c|}{Age±10} & Gender \\ \hline
ResNet-34 & 21M & 3.6B & 19.6\% & 65.5\% & \multicolumn{1}{c|}{87.1\%} & 90.3\% \\ \hline
\end{tabular}
\end{center}
\label{tab:utkface}
\end{table}



\begin{table}
\caption{Respiratory sound classification accuracy and model complexity with and without taking the demographic information into account.}
\scriptsize
\setlength{\tabcolsep}{0.9pt}
\begin{center}
\begin{tabular}{c|cc|ccccc|c}
\hline
\multicolumn{3}{c|}{DCNN characteristics} & \multicolumn{5}{c|}{Sensitivity} & Accuracy \\ \hline
Model & \#params & FLOPs & URTI & Healthy & COPD & Bronchiec. & Bronchiol. & Test \\ \hline
\begin{tabular}[c]{@{}c@{}}EnvNet-like on Sound \\ w/o Demographic Info\end{tabular} & 320k & 0.194B & 21\% & 68\% & 96\% & 88\% & 4\% & 78\% \\ \hline
\begin{tabular}[c]{@{}c@{}}EnvNet-like on Sound \\ with Demographic Info\end{tabular} & 320k & 0.194B & 16\% & 72\% & 100\% & 88\% & 15\% & 83\% \\ \hline
\end{tabular}
\end{center}
\label{tab:icbhi_2}
\end{table}

\subsection{Auditory Classification}
\subsubsection{EnvNet for Respiratory Sound and Demographic Information}\hfill\\
We first conduct a set of experiments to explore the best DCNN configuration based on the EnvNet DCNN that achieves the highest accuracy. Then, we combined the audio dataset with the age groups they are recorded from as depicted in Fig. \ref{fig:proposed}. Table \ref{tab:icbhi_2} compares the two sets of experiments, indicating that the COPD and healthy conditions are diagnosed with higher accuracy and resulting in a total test accuracy increase by 5\% when the demographic information is taken into account. 

\section{Commercial Off-The-Shelf Device Deployment}
The framework is intended to be flexibly deployable for general-purpose devices where the developed ML models trained on the framework can be deployed onto processing machines that may range from front-end edge devices to back-end computer servers. Trading off between the computation complexity and the classification accuracy, trained ML models can be deployed to edge devices (e.g. a cell-phone, tablet) to process data locally if the information privacy is a concern, or otherwise to the cloud servers that can process data with more elaborate up-to-date models that yield higher quality metrics. 

All of the DCNN models are attributed to at least two hardware-level characteristics: the model size and the number of computer operations per inference, both of which are upper-bounded by the platform resources that they deploy to, or by the inference deadline. When putting all the components of the framework together, both the hardware resource constraints and the diagnosis latency should meet the application goals. Having set the batch-size equal to 1, the trained models obtained from the previous Section are deployed on two mobile CPUs including Denver (dual-core) and ARM-Cortex A57 (quad-core) as well as an embedded CPU+GPU implementation with different frequency settings. All of the settings were performed on the TX2 development board that provides precise on-board power measurement. Table \ref{tab:cots} summarizes the implementation, indicating that, provided a 5s frame of recording to the memory, the least power dissipating implementation (Denver with a low frequency) takes 10 seconds to classify one frame whereas the most energy-efficient implementation (CPU+GPU) dissipates approximately 10$\times$ more power to classify the same frame within 0.1 seconds.


\begin{table}[]
\centering
\caption{Deploying the EnvNet model to commercial off-the-shelf devices including a dual-core Denver CPU, a quad-core ARM A57 CPU, and a combination of ARM CPU + Pascal GPU from the NVIDIA TX2 board.}
\label{tab:cots}
\resizebox{0.47\textwidth}{!}{%
\begin{tabular}{c|c|c|c|c|c|c|c}
\hline
Configuration & \begin{tabular}[c]{@{}c@{}}CPU Freq.\\ (MHz)\end{tabular} & \begin{tabular}[c]{@{}c@{}}GPU Freq.\\ (MHz)\end{tabular} & \begin{tabular}[c]{@{}c@{}}Power\\ (mW)\end{tabular} & \begin{tabular}[c]{@{}c@{}}Latency\\ (S)\end{tabular} & \begin{tabular}[c]{@{}c@{}}Performance\\ (GFLOP/S)\end{tabular} & \begin{tabular}[c]{@{}c@{}}Energy\\ (J)\end{tabular} & \begin{tabular}[c]{@{}c@{}}Energy Eff\\ (GFLOPS/W)\end{tabular} \\ \hline
\multirow{2}{*}{Denver CPU} & 345 & - & 881 & 10.0 & 0.019 & 8.81 & 0.021 \\ \cline{2-8} 
 & 2035 & - & 3170 & 0.9 & 0.215 & 2.85 & 0.068 \\ \hline
\multirow{2}{*}{ARM A57 CPU} & 345 & - & 1168 & 3.7 & 0.052 & 4.32 & 0.045 \\ \cline{2-8} 
 & 2035 & - & 4425 & 0.6 & 0.322 & 2.66 & 0.073 \\ \hline
TX2 CPU+GPU & 2035 & 1300.5 & 9106 & 0.1 & 1.935 & 0.91 & 0.210 \\ \hline
\end{tabular}%
}
\end{table}

\section{Comparison}
The most related work to ours that has developed a DCNN on the same respiratory sound database is the work in \cite{tariq2019lung} that reports an overall accuracy of 97\%. The main difference between the two works is that our model uses additional information in tandem with the audio data and proposes a framework that suggests combining as much existing correlated information within the dataset as possible to rectify and increase the diagnosis accuracy. The other difference is that our selected dataset is semi-balanced among 5 classes of respiratory sounds recorded from one unique medical device that has been indistinguishably utilized for 61 subjects diagnosed with 7 out of 8 classes within the database, whereas the dataset selection in \cite{tariq2019lung} is excessively dominated with COPD recordings, a major portion of which, as depicted in Fig. \ref{fig:ICHBI}-D, are recorded by two medical devices that have been used to merely sample from COPD-diagnosed participants. Table \ref{tab:comparison} provides a comparison and a summary of the total number of augmented samples per class within the two works.

\begin{table}
\caption{Comparison to the Related Work}
\scriptsize
\setlength{\tabcolsep}{2.1pt}
\begin{center}
\begin{tabular}{c|c|c|c|c|c|c|c|c|c}
\hline
\multirow{2}{*}{\begin{tabular}[c]{@{}c@{}}Related\\ Work\end{tabular}} & \multicolumn{8}{c|}{\#Augmented Audio Samples within the Dataset} & \multirow{2}{*}{\begin{tabular}[c]{@{}c@{}}Test\\ Accuracy\end{tabular}} \\ \cline{2-9}
 & URTI & Healthy & COPD & Bronchiec. & Bronchiol. & Asthma & LRTI & Pneum. &  \\ \hline
{\cite{tariq2019lung}} & 403 & 455 & 10,205 & 377 & - & 13 & 26 & 481 & 97\% \\ \hline
This Work & 370 & 560 & 567 & 256 & 208 & - & - & - & 83\% \\ \hline
\end{tabular}
\end{center}
\label{tab:comparison}
\end{table}

\section{Conclusion}
In an attempt to exploit machine learning algorithms to classify respiratory problems, we proposed a framework that employs as much correlated information as a dataset provides and showed that with combining both auditory and demographic information for a selection of reasonably balanced dataset out of a publicly released respiratory sound database the diagnosis accuracy of the trained deep convolutional neural networks (DCNNs) increases by 5\%. Since the demographic data can be extracted and estimated using computer vision, we suggest using another DCNN that works in parallel to the auditory signal processing DCNN to estimate the demographic information of the subject under test. Lastly, we deploy our DCNN models on a dual-core Denver CPU, a quad-core ARM Cortex A57, and a heterogeneous implementation of CPU+GPU from the NVIDIA TX2 development board to measure hardware characteristics when deploying the model to an embedded device.

\bibliographystyle{ACM-Reference-Format}
\bibliography{epiDAMIK}


\begin{thebibliography}{23}


\ifx \showCODEN    \undefined \def \showCODEN     #1{\unskip}     \fi
\ifx \showDOI      \undefined \def \showDOI       #1{#1}\fi
\ifx \showISBNx    \undefined \def \showISBNx     #1{\unskip}     \fi
\ifx \showISBNxiii \undefined \def \showISBNxiii  #1{\unskip}     \fi
\ifx \showISSN     \undefined \def \showISSN      #1{\unskip}     \fi
\ifx \showLCCN     \undefined \def \showLCCN      #1{\unskip}     \fi
\ifx \shownote     \undefined \def \shownote      #1{#1}          \fi
\ifx \showarticletitle \undefined \def \showarticletitle #1{#1}   \fi
\ifx \showURL      \undefined \def \showURL       {\relax}        \fi
\providecommand\bibfield[2]{#2}
\providecommand\bibinfo[2]{#2}
\providecommand\natexlab[1]{#1}
\providecommand\showeprint[2][]{arXiv:#2}

\bibitem[\protect\citeauthoryear{World Health Organization}{who}{2020}]%
        {who2016}
 \bibinfo{year}{2018 (accessed June, 2020)}\natexlab{}.
\newblock \showarticletitle{The Top 10 Causes of Death}. World Health
  Organization.
\newblock
\urldef\tempurl%
\url{https://www.who.int/news-room/fact-sheets/detail/the-top-10-causes-of-death}
\showURL{%
\tempurl}


\bibitem[\protect\citeauthoryear{Acharya and Basu}{Acharya and Basu}{2020}]%
        {acharya2020deep}
\bibfield{author}{\bibinfo{person}{Jyotibdha Acharya} {and}
  \bibinfo{person}{Arindam Basu}.} \bibinfo{year}{2020}\natexlab{}.
\newblock \showarticletitle{Deep Neural Network for Respiratory Sound
  Classification in Wearable Devices Enabled by Patient Specific Model Tuning}.
\newblock \bibinfo{journal}{\emph{IEEE transactions on biomedical circuits and
  systems}} \bibinfo{volume}{14}, \bibinfo{number}{3} (\bibinfo{year}{2020}),
  \bibinfo{pages}{535--544}.
\newblock


\bibitem[\protect\citeauthoryear{Amoh and Odame}{Amoh and Odame}{2016}]%
        {amoh2016deep}
\bibfield{author}{\bibinfo{person}{Justice Amoh} {and} \bibinfo{person}{Kofi
  Odame}.} \bibinfo{year}{2016}\natexlab{}.
\newblock \showarticletitle{Deep neural networks for identifying cough sounds}.
\newblock \bibinfo{journal}{\emph{IEEE transactions on biomedical circuits and
  systems}} \bibinfo{volume}{10}, \bibinfo{number}{5} (\bibinfo{year}{2016}),
  \bibinfo{pages}{1003--1011}.
\newblock


\bibitem[\protect\citeauthoryear{Aykanat, K{\i}l{\i}{\c{c}}, Kurt, and
  Saryal}{Aykanat et~al\mbox{.}}{2017}]%
        {aykanat2017classification}
\bibfield{author}{\bibinfo{person}{Murat Aykanat}, \bibinfo{person}{{\"O}zkan
  K{\i}l{\i}{\c{c}}}, \bibinfo{person}{Bahar Kurt}, {and}
  \bibinfo{person}{Sevgi Saryal}.} \bibinfo{year}{2017}\natexlab{}.
\newblock \showarticletitle{Classification of lung sounds using convolutional
  neural networks}.
\newblock \bibinfo{journal}{\emph{EURASIP Journal on Image and Video
  Processing}} \bibinfo{volume}{2017}, \bibinfo{number}{1}
  (\bibinfo{year}{2017}), \bibinfo{pages}{65}.
\newblock


\bibitem[\protect\citeauthoryear{Demir, Sengur, and Bajaj}{Demir
  et~al\mbox{.}}{2020}]%
        {demir2020convolutional}
\bibfield{author}{\bibinfo{person}{Fatih Demir}, \bibinfo{person}{Abdulkadir
  Sengur}, {and} \bibinfo{person}{Varun Bajaj}.}
  \bibinfo{year}{2020}\natexlab{}.
\newblock \showarticletitle{Convolutional neural networks based efficient
  approach for classification of lung diseases}.
\newblock \bibinfo{journal}{\emph{Health Information Science and Systems}}
  \bibinfo{volume}{8}, \bibinfo{number}{1} (\bibinfo{year}{2020}),
  \bibinfo{pages}{4}.
\newblock


\bibitem[\protect\citeauthoryear{He, Zhang, Ren, and Sun}{He
  et~al\mbox{.}}{2016}]%
        {he2016deep}
\bibfield{author}{\bibinfo{person}{Kaiming He}, \bibinfo{person}{Xiangyu
  Zhang}, \bibinfo{person}{Shaoqing Ren}, {and} \bibinfo{person}{Jian Sun}.}
  \bibinfo{year}{2016}\natexlab{}.
\newblock \showarticletitle{Deep residual learning for image recognition}. In
  \bibinfo{booktitle}{\emph{Proceedings of the IEEE conference on computer
  vision and pattern recognition}}. \bibinfo{pages}{770--778}.
\newblock


\bibitem[\protect\citeauthoryear{Imran, Posokhova, Qureshi, Masood, Riaz, Ali,
  John, and Nabeel}{Imran et~al\mbox{.}}{2020}]%
        {imran2020ai4covid}
\bibfield{author}{\bibinfo{person}{Ali Imran}, \bibinfo{person}{Iryna
  Posokhova}, \bibinfo{person}{Haneya~N Qureshi}, \bibinfo{person}{Usama
  Masood}, \bibinfo{person}{Sajid Riaz}, \bibinfo{person}{Kamran Ali},
  \bibinfo{person}{Charles~N John}, {and} \bibinfo{person}{Muhammad Nabeel}.}
  \bibinfo{year}{2020}\natexlab{}.
\newblock \showarticletitle{AI4COVID-19: AI enabled preliminary diagnosis for
  COVID-19 from cough samples via an app}.
\newblock \bibinfo{journal}{\emph{arXiv preprint arXiv:2004.01275}}
  (\bibinfo{year}{2020}).
\newblock


\bibitem[\protect\citeauthoryear{K{\"a}rkk{\"a}inen and Joo}{K{\"a}rkk{\"a}inen
  and Joo}{2019}]%
        {karkkainen2019fairface}
\bibfield{author}{\bibinfo{person}{Kimmo K{\"a}rkk{\"a}inen} {and}
  \bibinfo{person}{Jungseock Joo}.} \bibinfo{year}{2019}\natexlab{}.
\newblock \showarticletitle{FairFace: Face Attribute Dataset for Balanced Race,
  Gender, and Age}.
\newblock \bibinfo{journal}{\emph{arXiv preprint arXiv:1908.04913}}
  (\bibinfo{year}{2019}).
\newblock


\bibitem[\protect\citeauthoryear{Lee, Hu, Chen, Huang, and Hsueh}{Lee
  et~al\mbox{.}}{2020}]%
        {lee2020children}
\bibfield{author}{\bibinfo{person}{Ping-Ing Lee}, \bibinfo{person}{Ya-Li Hu},
  \bibinfo{person}{Po-Yen Chen}, \bibinfo{person}{Yhu-Chering Huang}, {and}
  \bibinfo{person}{Po-Ren Hsueh}.} \bibinfo{year}{2020}\natexlab{}.
\newblock \showarticletitle{Are children less susceptible to COVID-19?}
\newblock \bibinfo{journal}{\emph{Journal of Microbiology, Immunology, and
  Infection}} (\bibinfo{year}{2020}).
\newblock


\bibitem[\protect\citeauthoryear{Liu, Chen, Lin, and Han}{Liu
  et~al\mbox{.}}{2020}]%
        {liu2020clinical}
\bibfield{author}{\bibinfo{person}{Kai Liu}, \bibinfo{person}{Ying Chen},
  \bibinfo{person}{Ruzheng Lin}, {and} \bibinfo{person}{Kunyuan Han}.}
  \bibinfo{year}{2020}\natexlab{}.
\newblock \showarticletitle{Clinical features of COVID-19 in elderly patients:
  A comparison with young and middle-aged patients}.
\newblock \bibinfo{journal}{\emph{Journal of Infection}}
  (\bibinfo{year}{2020}).
\newblock


\bibitem[\protect\citeauthoryear{Liu, Cai, Zhang, and Hu}{Liu
  et~al\mbox{.}}{2019}]%
        {liu2019detection}
\bibfield{author}{\bibinfo{person}{Renyu Liu}, \bibinfo{person}{Shengsheng
  Cai}, \bibinfo{person}{Kexin Zhang}, {and} \bibinfo{person}{Nan Hu}.}
  \bibinfo{year}{2019}\natexlab{}.
\newblock \showarticletitle{Detection of Adventitious Respiratory Sounds based
  on Convolutional Neural Network}. In \bibinfo{booktitle}{\emph{2019
  International Conference on Intelligent Informatics and Biomedical Sciences
  (ICIIBMS)}}. IEEE, \bibinfo{pages}{298--303}.
\newblock


\bibitem[\protect\citeauthoryear{Ma, Xu, Yu, Zhang, Li, Zhao, and Wang}{Ma
  et~al\mbox{.}}{2019}]%
        {ma2019lungbrn}
\bibfield{author}{\bibinfo{person}{Yi Ma}, \bibinfo{person}{Xinzi Xu},
  \bibinfo{person}{Qing Yu}, \bibinfo{person}{Yuhang Zhang},
  \bibinfo{person}{Yongfu Li}, \bibinfo{person}{Jian Zhao}, {and}
  \bibinfo{person}{Guoxing Wang}.} \bibinfo{year}{2019}\natexlab{}.
\newblock \showarticletitle{LungBRN: A Smart Digital Stethoscope for Detecting
  Respiratory Disease Using bi-ResNet Deep Learning Algorithm}. In
  \bibinfo{booktitle}{\emph{2019 IEEE Biomedical Circuits and Systems
  Conference (BioCAS)}}. IEEE, \bibinfo{pages}{1--4}.
\newblock


\bibitem[\protect\citeauthoryear{Nakano, Furukawa, and Tanigawa}{Nakano
  et~al\mbox{.}}{2019}]%
        {nakano2019tracheal}
\bibfield{author}{\bibinfo{person}{Hiroshi Nakano}, \bibinfo{person}{Tomokazu
  Furukawa}, {and} \bibinfo{person}{Takeshi Tanigawa}.}
  \bibinfo{year}{2019}\natexlab{}.
\newblock \showarticletitle{Tracheal sound analysis using a deep neural network
  to detect sleep apnea}.
\newblock \bibinfo{journal}{\emph{Journal of Clinical Sleep Medicine}}
  \bibinfo{volume}{15}, \bibinfo{number}{8} (\bibinfo{year}{2019}),
  \bibinfo{pages}{1125--1133}.
\newblock


\bibitem[\protect\citeauthoryear{Perna}{Perna}{2018}]%
        {perna2018convolutional}
\bibfield{author}{\bibinfo{person}{Diego Perna}.}
  \bibinfo{year}{2018}\natexlab{}.
\newblock \showarticletitle{Convolutional neural networks learning from
  respiratory data}. In \bibinfo{booktitle}{\emph{2018 IEEE International
  Conference on Bioinformatics and Biomedicine (BIBM)}}. IEEE,
  \bibinfo{pages}{2109--2113}.
\newblock


\bibitem[\protect\citeauthoryear{Perna and Tagarelli}{Perna and
  Tagarelli}{2019}]%
        {perna2019deep}
\bibfield{author}{\bibinfo{person}{Diego Perna} {and} \bibinfo{person}{Andrea
  Tagarelli}.} \bibinfo{year}{2019}\natexlab{}.
\newblock \showarticletitle{Deep auscultation: Predicting respiratory anomalies
  and diseases via recurrent neural networks}. In
  \bibinfo{booktitle}{\emph{2019 IEEE 32nd International Symposium on
  Computer-Based Medical Systems (CBMS)}}. IEEE, \bibinfo{pages}{50--55}.
\newblock


\bibitem[\protect\citeauthoryear{Pham, McLoughlin, Phan, Tran, Nguyen, and
  Palaniappan}{Pham et~al\mbox{.}}{2020}]%
        {pham2020robust}
\bibfield{author}{\bibinfo{person}{Lam Pham}, \bibinfo{person}{Ian McLoughlin},
  \bibinfo{person}{Huy Phan}, \bibinfo{person}{Minh Tran},
  \bibinfo{person}{Truc Nguyen}, {and} \bibinfo{person}{Ramaswamy
  Palaniappan}.} \bibinfo{year}{2020}\natexlab{}.
\newblock \showarticletitle{Robust Deep Learning Framework For Predicting
  Respiratory Anomalies and Diseases}.
\newblock \bibinfo{journal}{\emph{arXiv preprint arXiv:2002.03894}}
  (\bibinfo{year}{2020}).
\newblock


\bibitem[\protect\citeauthoryear{Rocha, Filos, Mendes, Vogiatzis, Perantoni,
  Kaimakamis, Natsiavas, Oliveira, J{\'a}come, Marques, et~al\mbox{.}}{Rocha
  et~al\mbox{.}}{2017}]%
        {rocha2017alpha}
\bibfield{author}{\bibinfo{person}{BM Rocha}, \bibinfo{person}{D Filos},
  \bibinfo{person}{L Mendes}, \bibinfo{person}{I Vogiatzis}, \bibinfo{person}{E
  Perantoni}, \bibinfo{person}{E Kaimakamis}, \bibinfo{person}{P Natsiavas},
  \bibinfo{person}{A Oliveira}, \bibinfo{person}{C J{\'a}come},
  \bibinfo{person}{A Marques}, {et~al\mbox{.}}}
  \bibinfo{year}{2017}\natexlab{}.
\newblock \showarticletitle{A respiratory sound database for the development of
  automated classification}. In \bibinfo{booktitle}{\emph{International
  Conference on Biomedical and Health Informatics}}. Springer,
  \bibinfo{pages}{33--37}.
\newblock


\bibitem[\protect\citeauthoryear{Rocha, Filos, Mendes, Serbes, Ulukaya, Kahya,
  Jakovljevic, Turukalo, Vogiatzis, Perantoni, et~al\mbox{.}}{Rocha
  et~al\mbox{.}}{2019}]%
        {rocha2019open}
\bibfield{author}{\bibinfo{person}{Bruno~M Rocha}, \bibinfo{person}{Dimitris
  Filos}, \bibinfo{person}{Lu{\'\i}s Mendes}, \bibinfo{person}{Gorkem Serbes},
  \bibinfo{person}{Sezer Ulukaya}, \bibinfo{person}{Yasemin~P Kahya},
  \bibinfo{person}{Nik{\v{s}}a Jakovljevic}, \bibinfo{person}{Tatjana~L
  Turukalo}, \bibinfo{person}{Ioannis~M Vogiatzis}, \bibinfo{person}{Eleni
  Perantoni}, {et~al\mbox{.}}} \bibinfo{year}{2019}\natexlab{}.
\newblock \showarticletitle{An open access database for the evaluation of
  respiratory sound classification algorithms}.
\newblock \bibinfo{journal}{\emph{Physiological measurement}}
  \bibinfo{volume}{40}, \bibinfo{number}{3} (\bibinfo{year}{2019}),
  \bibinfo{pages}{035001}.
\newblock


\bibitem[\protect\citeauthoryear{Ryu, Park, and Shin}{Ryu
  et~al\mbox{.}}{2016}]%
        {ryu2016classification}
\bibfield{author}{\bibinfo{person}{Heechang Ryu}, \bibinfo{person}{Jinkyoo
  Park}, {and} \bibinfo{person}{Hayong Shin}.} \bibinfo{year}{2016}\natexlab{}.
\newblock \showarticletitle{Classification of heart sound recordings using
  convolution neural network}. In \bibinfo{booktitle}{\emph{2016 Computing in
  Cardiology Conference (CinC)}}. IEEE, \bibinfo{pages}{1153--1156}.
\newblock


\bibitem[\protect\citeauthoryear{Tariq, Shah, and Lee}{Tariq
  et~al\mbox{.}}{2019}]%
        {tariq2019lung}
\bibfield{author}{\bibinfo{person}{Zeenat Tariq},
  \bibinfo{person}{Sayed~Khushal Shah}, {and} \bibinfo{person}{Yugyung Lee}.}
  \bibinfo{year}{2019}\natexlab{}.
\newblock \showarticletitle{Lung Disease Classification using Deep
  Convolutional Neural Network}. In \bibinfo{booktitle}{\emph{2019 IEEE
  International Conference on Bioinformatics and Biomedicine (BIBM)}}. IEEE,
  \bibinfo{pages}{732--735}.
\newblock


\bibitem[\protect\citeauthoryear{Tokozume and Harada}{Tokozume and
  Harada}{2017}]%
        {tokozume2017learning}
\bibfield{author}{\bibinfo{person}{Yuji Tokozume} {and}
  \bibinfo{person}{Tatsuya Harada}.} \bibinfo{year}{2017}\natexlab{}.
\newblock \showarticletitle{Learning environmental sounds with end-to-end
  convolutional neural network}. In \bibinfo{booktitle}{\emph{2017 IEEE
  International Conference on Acoustics, Speech and Signal Processing
  (ICASSP)}}. IEEE, \bibinfo{pages}{2721--2725}.
\newblock


\bibitem[\protect\citeauthoryear{Zhang, Song, and Qi}{Zhang
  et~al\mbox{.}}{2017}]%
        {zhifei2017cvpr}
\bibfield{author}{\bibinfo{person}{Zhifei Zhang}, \bibinfo{person}{Yang Song},
  {and} \bibinfo{person}{Hairong Qi}.} \bibinfo{year}{2017}\natexlab{}.
\newblock \showarticletitle{Age progression/regression by conditional
  adversarial autoencoder}. In \bibinfo{booktitle}{\emph{Proceedings of the
  IEEE conference on computer vision and pattern recognition}}.
  \bibinfo{pages}{5810--5818}.
\newblock


\bibitem[\protect\citeauthoryear{Zhao, Zhang, Li, Ma, Gu, Hou, Guo, Wu, and
  Bai}{Zhao et~al\mbox{.}}{2020}]%
        {zhao2020incidence}
\bibfield{author}{\bibinfo{person}{Xianxian Zhao}, \bibinfo{person}{Bili
  Zhang}, \bibinfo{person}{Pan Li}, \bibinfo{person}{Chaoqun Ma},
  \bibinfo{person}{Jiawei Gu}, \bibinfo{person}{Pan Hou},
  \bibinfo{person}{Zhifu Guo}, \bibinfo{person}{Hong Wu}, {and}
  \bibinfo{person}{Yuan Bai}.} \bibinfo{year}{2020}\natexlab{}.
\newblock \showarticletitle{Incidence, clinical characteristics and prognostic
  factor of patients with COVID-19: a systematic review and meta-analysis}.
\newblock \bibinfo{journal}{\emph{MedRxiv}} (\bibinfo{year}{2020}).
\newblock


\end{thebibliography}
\end{document}